# Brain MRI-based 3D Convolutional Neural Networks for Classification of Schizophrenia and Controls *

Mengjiao Hu, Kang Sim, Juan Helen Zhou, Xudong Jiang, *Senior Member, IEEE*, Cuntai Guan, *Fellow, IEEE*

*Abstract*— Convolutional Neural Network (CNN) has been successfully applied on classification of both natural images and medical images but not yet been applied to differentiating patients with schizophrenia from healthy controls. Given the subtle, mixed, and sparsely distributed brain atrophy patterns of schizophrenia, the capability of automatic feature learning makes CNN a powerful tool for classifying schizophrenia from controls as it removes the subjectivity in selecting relevant spatial features. To examine the feasibility of applying CNN to classification of schizophrenia and controls based on structural Magnetic Resonance Imaging (MRI), we built 3D CNN models with different architectures and compared their performance with a handcrafted feature-based machine learning approach. Support vector machine (SVM) was used as classifier and Voxel-based Morphometry (VBM) was used as feature for handcrafted feature-based machine learning. 3D CNN models with sequential architecture, inception module and residual module were trained from scratch. CNN models achieved higher cross-validation accuracy than handcrafted feature-based machine learning. Moreover, testing on an independent dataset, 3D CNN models greatly outperformed handcrafted feature-based machine learning. This study underscored the potential of CNN for identifying patients with schizophrenia using 3D brain MR images and paved the way for imaging-based individual-level diagnosis and prognosis in psychiatric disorders.

## I. INTRODUCTION

Schizophrenia is a severe chronic mental disorder that causes great burdens to patients' families and society. Yet, diagnosis of schizophrenia is typically based on interview and clinical symptoms, which could be challenged by the complex and heterogeneous symptoms as well as many confounders such as the race, gender, and medication effects[1]–[3]. Early and accurate diagnosis of schizophrenia could facilitate treatment planning and improve the outcome of the illness. Instead of classical group-level comparisons, machine learning based on objective biomarkers such as medical imaging could help to establish a psychiatric diagnosis at the individual level.

Indeed, previous work on computer-aided classification of schizophrenia patients and controls using brain structural magnetic resonance imaging (MRI) mainly involve handcrafted feature-based machine learning approaches such as support vector machine (SVM), random forest (RF), and logistic regression (LR). MRI-based features include region- or voxel-specific cortical thickness and gray matter volume density in the brain [4]–[9]. However, many studies suffer from poor replicability and generalizability due to small sample sizes and hyper parameter settings. The inconsistent, subtle and sparsely distributed brain atrophy pattern in schizophrenia also limits the performance and generalizability of handcrafted feature-based machine learning as the selected features might not be reliable and hence imposes restrictions on the application of computer-aided diagnosis in real world. Handcrafted feature-based machine learning also lacks the ability to extract new features to make biological inferences [4], [6].

As a data-driven method, CNN is capable of automatic feature learning which removes the subjectivity in selecting relevant spatial features. This is especially important for psychiatric disorders like schizophrenia which has subtle and sparsely distributed brain alterations. Deep model architecture with non-linear layers also allows mapping of complicated data pattern [4], [10]. CNN has been successfully applied on medical images for classification of brain tumor, lung nodule, Alzheimer's disease and so on [10]–[14], but has not been applied to differentiating schizophrenia and controls. Therefore, this study aims to examine the feasibility of applying CNN to classification of schizophrenia patients and healthy controls based on structural MRI.

3D CNN models were proposed for classification of 3D images as they could fully utilize the context information of 3D images. However, development of 3D CNN were limited to simple architecture and small image size due to the high computational cost. Shallow 3D CNN models were proposed for classification of lung nodule and pulmonary [14], [15]. Small patches were utilized for quantification of perivascular spaces and classification of Alzheimer's disease [12], [16]. Meanwhile, multi-channel input is widely used in shallow networks to provide additional information for classification [11], [14].

Here, we developed 3D CNN models with different architectures and different depths to classify schizophrenia patients and healthy controls. Both cross-validation accuracy and testing accuracy on an independent dataset of 3D CNN models greatly outperformed handcrafted feature-based machine learning, which highlighted the potential of CNN for differentiating patients with schizophrenia from controls based on 3D structural MRI. This study paved the way for imaging-based individual-level diagnosis and prognosis in psychiatric disorders.

*Research supported by Institute of Health Technology, Nanyang Technological University, Singapore, Singapore

M. Hu, C. Guan and X. Jiang are with Nanyang Technological University, Singapore, Singapore (email: hume0004@e.ntu.edu.sg)

K. Sim is with Institute of Mental Health, Singapore, Singapore
M. Hu and J. Zhou are with National University of Singapore, Singapore, Singapore and Duke-NUS Medical School, Singapore, Singapore

## II. METHODS

### A. Datasets

Two independent structural MRI datasets of schizophrenia and controls were used in this study. The Northwestern University Schizophrenia Data and Software Tool (NUSDAST) is a repository of schizophrenia neuroimaging data collected from over 450 schizophrenia patients and healthy controls [17]. 141 schizophrenia patients and 134 healthy controls from this public dataset were included in training set after quality control. A similar dataset of 148 schizophrenia and 76 healthy controls from Institute of Mental Health (IMH), Singapore, were used as an additional testing set [18], [19] (Table I).

TABLE I. Demographic data for NUSDAST and IMH datasets

|  | NUSDAST | | IMH | |
| --- | --- | --- | --- | --- |
|  | SZ | HC | SZ | HC |
| N | 141 | 134 | 148 | 76 |
| AGE MEAN±STD | 35.1 ±12.8 | 32.9 ±14.0 | 32.7 ±9.0 | 31.3 ±9.8 |
| GENDER M/F | 90/51 | 72/62 | 102/46 | 47/29 |
| RESOLUTION | $1mm^3$ | | $1mm^3$ | |
| ORIGINAL DIMENSION | 180x256x256 | | 256x256x180 | |

(SZ-Schizophrenia Patients, HC-Healthy Controls, STD –standard deviation)

### B. Preprocessing

Preprocessing of structural MRI data (imaging parameters in Table I) was performed using CAT12 toolkit for voxel-wise estimation of the local amount of gray matter (GM), white matter (WM) and cerebrospinal fluid (CSF) compartment [20]. GM, WM and CSF probability maps were generated after skull striping, registration to standard space using MNI152 template, tissue segmentation and bias correction. For computational efficiency, all the resulting probability maps in the standard space were down-sampled from 121x145x121 to 61x73x61.

### C. Handcrafted Feature-based Machine Learning

Linear and non-linear SVM were employed to classify VBM features. GM, WM and CSF probability maps were flattened as feature vectors and feature reduction was completed by principal components analysis (PCA). Feature vectors were sent to linear and non-linear SVM for classification. Hyper-parameters for non-linear SVM were selected via grid search.

### D. 3D CNN Models

Typical CNN models consist of convolutional layers, pooling layers, and fully connected layers in sequential and backpropagation to learn multi-level features. Advanced architectures interconnect the layers and form modules with more complicated topologies. In this study, the following three types of 3D CNN model architectures with different depths were explored.

#### 1) Sequential Models

Sequential models followed the typical CNN architecture with convolutional layers, pooling layers, and fully connected layers in sequential. The convolutional kernel and pooling kernel were set with dimension 3x3x3 as selected using grid search. The feature map was flattened and connected to a fully connected layer with 128 neurons. Output was obtained by sigmoid function.

Three sequential models with different numbers of layers were trained and tested. With different depths, Sequential_1 (Seq_1) has structure Conv+Maxpooling+FC; Sequential_2 (Seq_2) has structure 2(Cov+Maxpooling)+FC; Sequential_3 (Seq_3) has structure 3(Conv+Maxpooling)+FC (Figure 1).

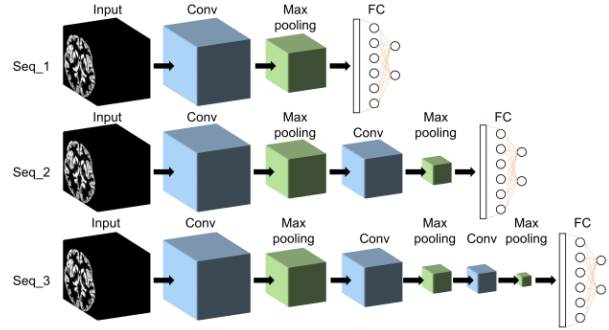

Figure 1. Architectures of Sequential Models

#### 2) Inception Models

Inspired by the GoogLeNet [21], 3D inception module was utilized in inception models. Inception module divides the network into multiple branches with different convolutional kernels. Figure 2 shows the structure of inception module and two models with different depths.

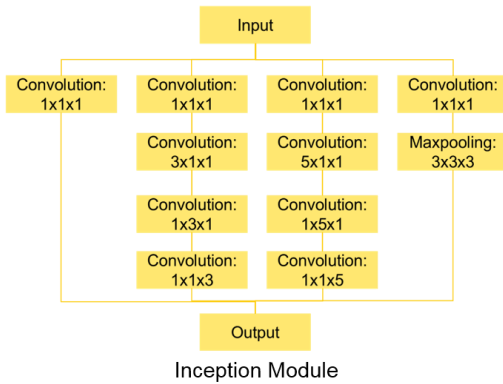
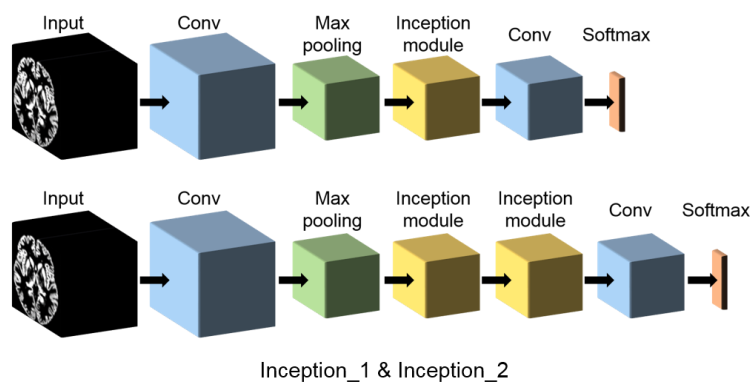

Figure 2. Architectures of Inception Module and Inception Models

3) Inception_resnet Models

Inspired by the residual module [22], Inception_resnet models combined inception architecture and residual module to utilize the information from previous layers. Same arterial structure as inception models were utilized with an extra connection which adds output from previous layer and output from inception module together. Two models with different depths were implemented (Figure 3).

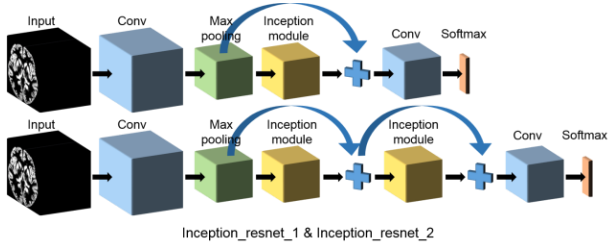

Figure 3. Architectures of Inception_resnet Models

*E. Multi-channel 3D Image Input*

To provide additional information for classification, GM, WM and CSF probability maps were used as multi-inputs for models. GM, WM and CSF probability maps went through feature reduction separately and then were concatenated together as the input feature vector in handcrafted feature-based machine learning approach. For 3D CNN models, GM, WM and CSF probability maps were sent to three parallel network branches and the outputs were concatenated together before feeding into the fully-connected layer or final convolutional layer.

*E. Nested Cross-validation and Ensemble Approach*

Nested cross-validation was utilized for hyperparameter selection and testing results obtaining. Testing results were obtained through outer 5-fold cross validation which averages the test error over multiple train–test splits. The testing results for 3D CNN models were reported as the average of 10 repeats to reduce randomness generated from training process.

After nested cross validation, 5 classifiers were trained in the outer loop for handcrafted feature-based machine learning. Independent testing results on IMH dataset were determined by voting among 5 predictions from the 5 models trained. For 3D CNN models, 10 repeats of the outer 5-fold cross-validation were performed to reduce the randomness. 5 models from the repeat with the highest accuracy out of 10 repeats were used as testing models. Similarly, the independent testing result on IMH dataset was determined by voting among 5 predictions. ROC-AUC was obtained by average correct votes.

### III. RESULTS

*A. Cross-validation Results on NUSDAST Dataset*

Accuracy, sensitivity, specificity and ROC-AUC obtained from nested cross-validation were reported for linear and non-linear SVM. Average of 10 repeats were reported for 3D CNN models. 3D CNN models achieved higher accuracy than handcrafted feature-based machine learning approaches in overall (Table II). The highest accuracy was obtained by Inception_resnet_1 at 79.27%

TABLE II. Cross-validation Results on NUSDAST Dataset

| | acc | sp | se | AUC |
|---|---|---|---|---|
| *Handcrafted Feature-Based Machine Learning* | | | | |
| **Linear SVM** | 60.77% | 60.48% | 61.08% | 0.646 |
| **Non-Linear SVM** | **69.15%** | **70.20%** | **68.10%** | **0.741** |
| *3D CNN Models* | | | | |
| **Sequential_1** | 77.63% | 80.39% | 75.00% | 0.802 |
| **Sequential_2** | 76.50% | 75.91% | 77.07% | 0.779 |
| **Sequential_3** | 73.91% | 73.13% | 74.57% | 0.753 |
| **Inception_1** | 78.19% | 77.22% | 79.07% | 0.795 |
| **Inception_2** | 76.24% | 78.08% | 74.42% | 0.773 |
| **Inception_Resnet_1** | **79.27%** | **80.44%** | **78.15%** | **0.803** |
| **Inception_Resnet_2** | 78.76% | 81.54% | 76.10% | 0.803 |

(acc-accuracy, sp-specificity, se-sensitivity)

*B. Additional Testing Results on IMH Dataset*

Accuracy, sensitivity and specificity were reported for testing group with results obtained by ensemble approach and shown in Table III. The highest testing accuracy was also obtained by inception_resnet_1 model at 70.98%. Handcrafted feature-based machine learning approaches obtained very low accuracy. 3D CNN models with fully connected layers also have low performance. 3D CNN models with more complex architectures obtained higher accuracy as well as more balanced sensitivity and specificity.

TABLE III. Additional Testing Results on IMH Dataset

| | acc | sp | se | AUC |
|---|---|---|---|---|
| *Handcrafted Feature-Based Machine Learning* | | | | |
| **Linear SVM** | 54.46% | 52.63% | 55.41% | 0.576 |
| **Non-Linear SVM** | 58.48% | 50.00% | 62.84% | 0.575 |
| *3D CNN Models* | | | | |
| **Sequential_1** | 62.95% | 75.00% | 56.76% | 0.733 |
| **Sequential_2** | 54.46% | 93.42% | 34.46% | 0.718 |
| **Sequential_3** | 70.09% | 46.05% | 82.43% | 0.722 |
| **Inception_1** | 68.30% | 68.42% | 68.24% | 0.737 |
| **Inception_2** | 66.96% | 65.79% | 67.57% | 0.713 |
| **Inception_Resnet_1** | **70.98%** | **63.16%** | **75.00%** | **0.753** |
| **Inception_Resnet_2** | 66.96% | 72.37% | 64.19% | 0.609 |

(acc-accuracy, sp-specificity, se-sensitivity)

### IV. DISCUSSION

This study validated the use of CNN on classification of schizophrenia patients and controls based on structural MRI. 3D CNN models trained from scratch performed better than handcrafted feature-based machine learning, highlighting the potential of using CNN on classification of schizophrenia patients and controls. Unlike other diseases with obvious alterations on neuroimaging, the neuroanatomical alterations in schizophrenia tend to be subtle and sparsely distributed. The diagnosis of schizophrenia is typically based on interview and clinical symptoms instead of radiological results [2], [3]. Handcrafted feature-based machine learning approach requires feature extraction priori to classification, which limits the performance and generalizability of trained models as the features might be non-reliable and dependent on datasets. CNN as a data-driven method, automatically explores the anatomical context and learns features for classification. Furthermore, 3D CNN models obtained much higher accuracy than handcrafted feature-based machine learning when models trained on NUSDAST dataset were tested on IMH dataset, indicating better generalizability of 3D CNN models. Even though our accuracy is not high enough for real world application, our results suggested that CNN is able to capture the subtle features in neuroimaging that is not visible to human eyes. Our results also highlight the potential of utilizing CNN for individual-level diagnosis of schizophrenia as the generalizability is validated. Moreover, to the best of our acknowledgement, this is the first study utilizing CNN on classification of schizophrenia patients and controls.

Comparing different 3D model architectures trained from scratch, the complex topologies such as inception module and residual module improved the accuracy. The improvement was also observed in another study although the residual module didn't over perform inception architecture in their study [14]. The number of parameters were greatly reduced by removing the fully connected layer. Removing the fully connected layer also controlled overfitting and thus greatly improved the testing accuracy. Models with more layers could not further improve the accuracy. A study had the same observation and concluded that using 3D CNN to classify lung nodules in CT does not need a very deep network [14]. Our assumption is that it could be due to the lost spatial information during registration to standard template and causing the features could not be very well distinguished by very deep networks.

The major limitation of this study is that the sample size is relatively small especially for network training. The limited sample size might affect the performance of the model and resulted in lower accuracy and generalizability.

## V. Conclusion

To classify patients with schizophrenia and healthy controls using 3D brain MR images, we developed 3D CNN models which greatly outperformed handcrafted feature-based machine learning. Our study underscored the potential of CNN for identifying patients with schizophrenia using 3D brain MR images and paved the way for imaging-based individual-level diagnosis and prognosis in psychiatric disorders.

## Acknowledgment

The research is supported by Institute of Health Technology, Nanyang Technological University, Singapore, Singapore.

## References

[1] A. P. Association, *Diagnostic and statistical manual of mental disorders (DSM-5®)*. American Psychiatric Pub, 2013.
[2] A. H. Fanous, R. L. Amdur, F. A. O'Neill, D. Walsh, and K. S. Kendler, "Concordance between chart review and structured interview assessments of schizophrenic symptoms," *Compr. Psychiatry*, vol. 53, no. 3, pp. 275–279, 2012.
[3] J. L. Kennedy, C. A. Altar, D. L. Taylor, I. Degtiar, and J. C. Hornberger, "The social and economic burden of treatment-resistant schizophrenia: A systematic literature review," *Int. Clin. Psychopharmacol.*, vol. 29, no. 2, pp. 63–76, 2014.
[4] M. R. Arbabshirani, S. Plis, J. Sui, and V. D. Calhoun, "Single subject prediction of brain disorders in neuroimaging: Promises and pitfalls," *Neuroimage*, vol. 145, pp. 137–165, 2017.
[5] J. Kambeitz *et al.*, "Detecting neuroimaging biomarkers for schizophrenia: A meta-analysis of multivariate pattern recognition studies," *Neuropsychopharmacology*, vol. 40, no. 7, pp. 1742–1751, 2015.
[6] J. L. Winterburn *et al.*, "Can we accurately classify schizophrenia patients from healthy controls using magnetic resonance imaging and machine learning? A multi-method and multi-dataset study," *Schizophr. Res.*, 2017.
[7] S. Borgwardt *et al.*, "Distinguishing prodromal from first-episode psychosis using neuroanatomical single-subject pattern recognition," *Schizophr. Bull.*, vol. 39, no. 5, pp. 1105–1114, 2013.
[8] M. Nieuwenhuis, N. E. M. van Haren, H. E. H. Pol, W. Cahn, R. S. Kahn, and H. G. Schnack, "Classification of schizophrenia patients and healthy controls from structural MRI scans in two large independent samples," *Neuroimage*, vol. 61, no. 3, pp. 606–612, 2012.
[9] R. Chin, A. X. You, F. Meng, J. Zhou, and K. Sim, "Recognition of Schizophrenia with Regularized Support Vector Machine and Sequential Region of Interest Selection using Structural Magnetic Resonance Imaging," *Sci. Rep.*, vol. 8, no. 1, pp. 1–10, 2018.
[10] J. G. Lee *et al.*, "Deep learning in medical imaging: General overview," *Korean J. Radiol.*, vol. 18, no. 4, pp. 570–584, 2017.
[11] Y. Yang *et al.*, "Glioma grading on conventional MR images: A deep learning study with transfer learning," *Front. Neurosci.*, vol. 12, no. NOV, pp. 1–10, 2018.
[12] E. H. Asl *et al.*, "Alzheimer's disease diagnostics by a 3D deeply supervised adaptable convolutional network," *Front. Biosci. - Landmark*, vol. 23, no. 3, pp. 584–596, 2018.
[13] T. Iizuka, M. Fukasawa, and M. Kameyama, "Deep-learning-based imaging-classification identified cingulate island sign in dementia with Lewy bodies," *Sci. Rep.*, vol. 9, no. 1, pp. 1–9, 2019.
[14] K. Liu and G. Kang, "Multiview convolutional neural networks for lung nodule classification," *Int. J. Imaging Syst. Technol.*, vol. 27, no. 1, pp. 12–22, 2017.
[15] Q. Dou, H. Chen, L. Yu, J. Qin, and P. A. Heng, "Multilevel Contextual 3-D CNNs for False Positive Reduction in Pulmonary Nodule Detection," *IEEE Trans. Biomed. Eng.*, vol. 64, no. 7, pp. 1558–1567, 2017.
[16] F. Dubost *et al.*, "Enlarged perivascular spaces in brain MRI: Automated quantification in four regions," *Neuroimage*, vol. 185, pp. 534–544, 2019.
[17] L. Wang *et al.*, "Northwestern University Schizophrenia Data and Software Tool (NUSDAST)," *Front. Neuroinform.*, vol. 7, no. November, pp. 1–13, 2013.
[18] N. F. Ho *et al.*, "Progression from selective to general involvement of hippocampal subfields in schizophrenia," *Mol. Psychiatry*, vol. 22, no. 1, pp. 142–152, 2017.
[19] N. F. Ho *et al.*, "Progression from selective to general involvement of hippocampal subfields in schizophrenia," *Mol. Psychiatry*, vol. 22, no. 1, p. 142, 2017.
[20] F. Kurth, C. Gaser, and E. Luders, "A 12-step user guide for analyzing voxel-wise gray matter asymmetries in statistical parametric mapping (SPM)," *Nat. Protoc.*, vol. 10, no. 2, pp. 293–304, 2015.
[21] C. Szegedy *et al.*, "Going deeper with convolutions," *Proc. IEEE Comput. Soc. Conf. Comput. Vis. Pattern Recognit.*, vol. 07-12-June, pp. 1–9, 2015.
[22] K. He, X. Zhang, S. Ren, and J. Sun, "Deep residual learning for image recognition," *Proc. IEEE Comput. Soc. Conf. Comput. Vis. Pattern Recognit.*, vol. 2016-Decem, pp. 770–778, 2016.